\documentclass[pmlr,twoside,tablecaptionbottom]{pgm-jmlr}
\usepackage{pgm-pmlr-dat}% sets volume number and conference name

\usepackage{bm}
\firstpageno{1}% will be changed by proceedings editors
\usepackage{verbatim}
\usepackage{colortbl}
\usepackage{pgf, tikz}
\usetikzlibrary{arrows,automata,fit}
\usepgflibrary{shapes.geometric}

\usepackage{subcaption}

\usepackage[utf8]{inputenc}
\inputencoding{utf8}

\usepackage{booktabs}

\title[Learning Staged Trees from Incomplete Data]{Learning Staged Trees from Incomplete Data}

\author{\Name{Jack Storror Carter} \Email{jack.carter@dima.unige.it}\\
\addr Dipartimento di Matematica, Universit\`{a} degli Studi di Genova, Genova, Italy \AND
\Name{Manuele Leonelli} \Email{manuele.leonelli@ie.edu}\\
   \addr School of Science and Technology, IE University, Madrid, Spain \AND
   \Name{Eva Riccomagno} \Email{riccomagno@dima.unige.it}\\
   \addr Dipartimento di Matematica, Universit\`{a} degli Studi di Genova, Genova, Italy \AND
   \Name{Gherardo Varando} \Email{gherardo.varando@uv.es}\\
   \addr Image Processing Laboratory, Universitat de Val\`{e}ncia, Val\`{e}ncia, Spain}

\begin{document}
\maketitle

\begin{abstract}
Staged trees are probabilistic graphical models capable of representing any class of non-symmetric independence via a coloring of its vertices. Several structural learning routines have been defined and implemented to learn staged trees from data, under the frequentist or Bayesian paradigm. They assume a data set has been observed fully and, in practice, observations with missing entries are either dropped or imputed before learning the model. Here, we introduce the first algorithms for staged trees that handle missingness within the learning of the model. To this end, we characterize the likelihood of staged tree models in the presence of missing data and discuss pseudo-likelihoods that approximate it.  A structural expectation-maximization algorithm estimating the model directly from the full likelihood is also implemented and evaluated. A computational experiment showcases the performance of the novel learning algorithms, demonstrating that it is feasible to account for different missingness patterns when learning staged trees.
\end{abstract}


\section{Introduction}
\label{sec:intro}

Staged trees are a class of probabilistic graphical models  explicitly created for modelling scenarios with asymmetric sample spaces, those that cannot simply be written as sample products, and asymmetric independence relations \citep{collazo2018chain,smith2008conditional}. The underlying structure of the sample space is depicted by an event tree \citep{shafer1996art}, while independences are visually and formally represented by a coloring (also called \textit{staging}) of the non-leaf vertices of the tree. Chain event graphs are an equivalent, more compact, graphical representation of staged trees obtained by a coalescence of the non-leaf vertices. \citet{smith2008conditional} demonstrated that every Bayesian network (BN) can be represented by a staged tree, while the reverse does not hold.  

Just as for BNs, there has been a growing interest in defining machine learning algorithms for model selection of staged trees from data (also called \textit{structural learning}). \citet{freeman2011bayesian} introduced the first model selection algorithm within the Bayesian paradigm and under the assumption of a complete and independent data set. Structural learning algorithms under the frequentist paradigm requiring the same assumptions then also began to appear \citep[e.g.][]{carli2022r,silander2013dynamic}.

However, as noticed by \citet{scutari2020bayesian} in the context of BNs, in practical applications the assumption of a complete data set of independent observations is rarely tenable. While the assumption of independence has been relaxed by developing dynamic versions of staged trees \citep[see e.g.][]{barclay2015dynamic}, the case of non-complete data has received less attention. \citet{barclay2014chain} considered missing values as an additional variable level, thus extending the underlying event tree. This approach allowed for the identification of the generating missingness mechanism. Conversely, \citet{yu2021causal} considered the case of a hidden variable in the context of system reliability. 

In this paper we introduce the first generic model selection techniques to learn staged trees  when some observations have missing values. We formally derive the form of the likelihood in the case of incomplete data and note that it does not entail closed-form estimators as for the complete data case. This observation motivated the two approaches proposed in this paper: the first approximates the likelihood by a simpler version, called a \textit{pseudo-likelihood}, whose parameters can be estimated in closed form; the second uses approximating algorithms targeting the full likelihood, for instance the expectation-maximization (EM) algorithm \citep{dempster1977maximum}.  We perform a simulation study using staged trees from the literature to investigate the performance of the introduced algorithms.

\section{Staged Trees}

A \textit{staged tree} \citep{smith2008conditional,collazo2018chain} is a probabilistic graphical model for a process consisting of a sequence of discrete events.  It combines a probability tree with an equivalence relation on its non-leaf vertices.  To construct a staged tree we begin with an \textit{event tree} $\mathcal{T} = (V,E)$ consisting of a vertex set $V$ and a directed edge set $E$.  Edges in $E$ are written as an ordered pair of vertices where the edge $e = (v,w)$ is directed from the vertex $v$ to the vertex $w$.  The vertex set contains a single \textit{root} vertex $v_0$ with no incoming edges and at least two outgoing edges, representing the start of the process, and a number of \textit{leaf} vertices which have a single incoming edge and no outgoing edges, representing the end of the process.  Every other vertex has exactly one incoming edge and at least two outgoing edges. All non-leaf vertices (including the root) are called \textit{situations} and represent a possible state at which the process can arrive. The children of a vertex are denoted $\mathrm{ch}(v) = \{ w \in V : (v,w) \in E \}$.  A root-to-leaf path is a sequence of vertices $(v_0,v_1,\dots,v_k)$ such that $v_0$ is the root, $v_k$ is a leaf and $v_i \in \mathrm{ch}(v_{i-1})$ for each $i=1,\dots,k$.  The set of all root-to-leaf paths is denoted $\Lambda$.

The edges in the event tree are labeled such that for each situation, the outgoing edge labels describe all possible events that can occur at the next stage of the process.  We denote the label of an edge $(v,w)$ by $\mathrm{lab}(v,w)$.  A situation combined with its outgoing edges is called a \textit{floret}.  For each floret in the graph, one can associate a probability distribution, called the \textit{transition probabilities}, representing the conditional probabilities of the subsequent event of the process.  The transition probability associated to the edge $(v,w)$ is denoted by $\theta_{v,w}$.  The transition probabilities for the floret at $v$ are denoted $\theta_v = (\theta_{v,w})_{w \in \mathrm{ch}(v)}$.  The set of all transition probabilities is written $\theta = (\theta_v)_{v \in S}$, where $S$ is the set of all situations.  One obtains a joint distribution for the whole process by assigning a probability distribution to all florets and then using the standard chain rule of probability \citep{gorgen2015differential,gorgen2018discovery}. Therefore, the probability of a root-to-leaf path $\lambda=(v_0,v_1,\dots,v_k)$ is

\begin{equation}
\label{eq:path}
\theta_\lambda=\prod_{i=1}^k\theta_{v_{i-1},v_i}.
\end{equation}

The most general statistical model (the saturated model) places no further constraints on the probability distributions at each situation.  However, a staged tree model restricts the space by assuming that some situations (whose florets are identical in terms of topology) have the same probability distribution.  When this is the case, the two situations are said to be in the same \textit{stage}.  That is, two situations $v_1,v_2$ are in the same stage if $\theta_{v_1}=\theta_{v_2}$.  This is represented graphically by colouring vertices according to which stage they are in. In this paper we further require that florets are identical in terms of  edge labels and that the equality $\theta_{v_1}=\theta_{v_2}$ matches the edge labels.

\begin{figure}
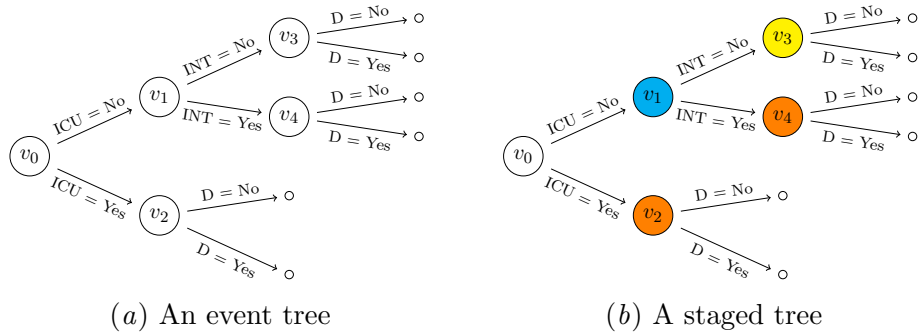

 % Caption and label go in the first argument and the figure contents
 % go in the second argument
 % the third argument is the figure; without subfigure just use \include... directly
\floatconts
  {fig:sts}
  {\caption{Examples of an event tree (left) and staged tree (right)}}
  {%
    \subfigure[An event tree]{\label{fig:compst}%
      \includeteximage[scale = 0.75]{comp}}%
    \qquad
\subfigure[A staged tree]{\label{fig:xorst}%
      \includeteximage[scale = 0.75]{xor}}
  }
\end{figure}

To illustrate this we consider a simplification of a staged tree from \citet{filigheddu2024using}. Patients with a specific condition arriving at a hospital might or might not enter the ICU. Patients who do not enter the ICU might be intubated, while those entering the ICU are by default intubated. Patients may pass away or not after a specific number of days. This scenario can be depicted by the event tree in Figure \ref{fig:compst}, which has five situations $v_0,\dots,v_4$ and six root-to-leaf paths. Now let's assume that non-intubated patients have the same probability of passing away irrespective of whether they entered the ICU or not. This assumption can be visually depicted by the coloring of the staged tree in Figure \ref{fig:xorst}.

%To illustrate staged trees we consider a simple biological example from \citet{gorgen2015differential}. A binary model is designed to explain a possible unfolding of the following events in a cell culture: a cell finds itself in a benign or hostile environment, the level of activity within this might be high or low, and if the environment is hostile then a cell might either survive or die. This scenario can be depicted by the event tree in Figure \ref{fig:compst}, which has five situations $v_0,\dots,v_4$ and six root-to-leaf paths. Let's now assume that a high or low level of activity is independent of the environment being hostile or benign and that whether or not a cell dies does not depend on its activity. These assumptions can be visually depicted by the coloring of the staged tree in Figure \ref{fig:xorst}.

\subsection{Learning Staged Trees from Complete Data}
A sample from an event tree $x = (x_1,\dots,x_k)$ is a sequence of events uniquely matching the edge labels in a root-to-leaf path of the tree.  That is, there is a unique root to leaf path $\lambda = (v_0,v_1,\dots,v_k)$ such that $\mathrm{lab}(v_{i-1},v_i) = x_i$ for each $i=1,\dots,k$.  A data set $D$ can therefore be summarised by the number of samples observed for each root-to-leaf path. We denote these counts by $n_\lambda$ for $\lambda \in \Lambda$.  The likelihood function can be written as 
\begin{equation}\label{eq:likelihood}
%l(\theta \mid D) = \sum_{\lambda \in \Lambda} n_\lambda \log(\theta_\lambda).
L(\theta \mid D) = \prod_{\lambda \in \Lambda}  {\theta_\lambda}^{n_\lambda}.
\end{equation}

%where $\theta_\lambda = \prod_{i=1}^k \theta_{v_{i-1},v_i}$ is the product of transition probabilities in the path.

Under standard assumptions over $D$, \citet{freeman2011bayesian} showed that Equations (\ref{eq:path}) and (\ref{eq:likelihood}) can be combined so that the staged tree likelihood factorizes over the stages of the model:
\begin{equation}
    L(\theta \mid D) = \prod_{s \in \mathcal{S}}\prod_{w\in\mathrm{ch}(v_s)}(\theta_{v_s,w})^{n_{v_s,w}},
    \label{eq:factstage}
\end{equation}
where $\mathcal{S}$ is the set of stages, $v_s$ is a representative vertex from the stage $s$ and $n_{v_s,w}$ is the number of observations in $D$ along the edge $(v_s,w)$ summed over all vertices $v_s \in s$. Because of the factorization in Equation (\ref{eq:factstage}), the transition probabilities $\theta_{v_s,w}$ can be easily estimated as the relative frequencies and the maximum likelihood estimator is $\hat{\theta}_{v_s,w}=n_{v_s,w}/\sum_{w'\in\mathrm{ch}(v_s)}n_{v_s,w'}$. In the Bayesian framework, independent conjugate Dirichlet priors can be given to each stage, resulting in a posterior parameter associated to $\theta_{v_s,w}$ equal to the sum of $n_{v_s,w}$ and its associated prior parameter. Thus, given a coloring of an event tree, it is straightforward to learn the transition probabilities from complete data.

%\begin{equation}
%    L(\theta \mid D) = \prod_{i=1}^U\prod_{j\in\mathrm{ch}(i)}\theta_{i,j}^{n_{i,j}},
%    \label{eq:factstage}
%\end{equation}
%where $U$ is the number of stages and $n_{i,j}$ is the number of observations in $D$ along the edge $(i,j)$. Because of the factorization in Equation (\ref{eq:factstage}), the transition probabilities $\theta_{i,j}$ can be easily estimated as the relative frequencies and the maximum likelihood estimator is $\hat{\theta}_{i,j}=n_{i,j}/\sum_{j\in\mathrm{ch}(i)}n_{i,j}$. In the Bayesian framework, independent conjugate Dirichlet priors can be given to each stage, giving as posterior mean of $\theta_{i,j}$ the sum of $n_{i,j}$ and its associated prior parameter. Thus, given a coloring of an event tree, it is straightforward to learn the transition probabilities under complete data.

There are two levels of model selection for staged trees. The first assumes that the underlying event tree is fixed meaning model selection is only over the coloring of the vertices, which are learned from data. Originally, this was performed exclusively using an agglomerative hierarchical clustering algorithm along with a Bayesian scoring function \citep{freeman2011bayesian}. This is implemented in the \texttt{cegpy} Python library \citep{walley2023cegpy}. Recently, learning algorithms using frequentist scores have been introduced and implemented in the \texttt{stagedtrees} R package \citep[see e.g.][]{leonelli2024structural,silander2013dynamic,varando2024staged}. The second level of model selection aims to also learn the underlying event tree $\mathcal{T}$: the ordering of the evolution of the process under study. Dynamic programming algorithms have been developed and implemented in \texttt{stagedtrees} for this task \citep{leonelli2023context,silander2013dynamic}. Learning the underlying event tree is critical when using observational data to infer causal relationships \citep{cowell2014causal}. 

\subsection{Staged Trees and Incomplete Data}

Incomplete data may appear in three different forms in staged trees: structural zeros, sampling zeros, and missing values.  Staged trees were originally defined to explicitly represent asymmetric processes which cannot be modeled by a product sample space, as usually assumed by BNs. Most often this consists of event trees which are not fully symmetric and with root-to-leaf paths of different lengths. Figure \ref{fig:compst} gives an example of structural zeros since patients who enter the ICU are always intubated - there is zero probability of not being intubated \citep[such trees are usually called non-$X$-compatible, see e.g. ][for an example]{leonelli2019sensitivity}. Outside of staged trees, structural zeros are generally difficult to model; one alternative approach to staged trees is to consider extensions of log-linear models for contingency tables \citep{klimova2016closure,klimova2022hierarchical}.

%In staged trees these are usually represented by missing edges in the underlying graphical representation. When structural zeros are present, all root-to-leaf paths do not have the same length. \color{red} I think they do for structural zeros if the data is otherwise stratified? It's more that different situations associated to the same variable do not necessarily have the same number of edges.  The difference in root to leaf path length comes from non-stratified examples. This makes me think of a further distinction for structural missingness if we think of the data in terms of variables
%{\color{green}{ER: The issue is thinking in terms of variables. Here I won't mention the distinction betwwen stratified and not s.}}:
%\begin{itemize}
%    \item Structural zero's where certain combinations of the variable values cannot be observed together i.e. certain paths in the event tree are probability 0.
%    \item Structurally missing values due to non-stratified data e.g. $X_2$ is only observed if $X_1 = x$.  This can either be considered directly in a non-stratified way (for example, when the value of $X_2$ really has no meaning if $X_1 \neq x$), or in a missing values way (maybe if $X_2$ always has some meaning but is less relevant when $X_1 \neq x$ - e.g. in a hospital, blood pressure is only measured for certain symptoms, but still has some meaning for all patients).
%\end{itemize}
%\color{black}

Sampling zeros occur when it is possible for a certain path to be observed, but it does not appear in the data. One of the first solutions was to eliminate those edges from the graph, thus reducing the space and time complexity of structural learning algorithms \citep{silander2013dynamic}. %This option is available in the \texttt{stagedtrees} R package through the input \texttt{join.unobserved}. 
Formally, all vertices with no observations are joined in the same stage and excluded from structural learning. Recently, \citet{carter2024staged} graphically visualized such an approach by adding a vertex to the equivalent chain event graph merging all unobserved vertices at any depth of the tree and dashing unobserved edges to clearly visualize paths with no data information. If, on the other hand, one wants to retain the full event tree, then the data has no information about the parameters associated to unobserved edges. In this case, everything is driven by the prior in a Bayesian approach \citep{freeman2011bayesian,hughes2022score} or by Laplace smoothing \citep{russell2016artificial}. An approach that aims to avoid observed zeros is to start the model search from a staged tree in which certain vertices are already merged in the same stage \citep{carter2024staged,leonelli2022highly,leonelli2024learning}.

%In some simple cases, \texttt{stagedtrees} \citep{carli2022r} can indirectly deal with such a situation. Conversely, \texttt{cegpy} \citep{walley2023cegpy} allows for more flexibility, at the cost of a trickier definition of the underlying model.

%\subsection{Observed zeros}

%This is the case when at a certain depth of the tree there are no observations along an edge. However, such observations can exist, it is simply we did not observe them.

%One of the first solutions was to eliminate those edges from the graph, thus reducing the space and time complexity of structural learning algorithms \citep{silander2013dynamic}. This option is available in the \texttt{stagedtrees} R package through the input \texttt{join.unobserved}. Formally, all vertices with no observations are joined in the same stage and excluded from structural learning.  This is also the default approach in the \texttt{cegpy} Python package where the event tree is constructed only using paths that are observed in the data.  If one wishes to include the zero observation paths then they must be added manually.

%Recently, \citet{carter2024staged} graphically visualized such an approach by adding a novel vertex to a CEG merging all unobserved vertices at any depth of the tree and dashing such edges to clearly identify them \citep[see also][]{filigheddu2024using}.

The more traditional form of incomplete data is when certain values from a sampled root-to-leaf path are hidden or not available to the user. 
The way in which the data is missing can be characterised as missing completely at random (MCAR), missing at random (MAR) or missing not at random (MNAR) \citep{little2019statistical,rubin1976inference}.  %STs have successfully been used to represent these different forms of missingness and can be used to learn which of the three missingness mechanisms the data is likely to follow \cite{barclay2014chain}.
When data is MNAR, it is often important to directly model the effect that missing data has on the probability distribution of non-missing data.  This can be easily incorporated into a staged tree model by including additional edges in the event tree that correspond to missing values \citep{barclay2014chain}, so to estimate both the probability of the data being missing and transition probabilities that condition on the past event being missing.  % The \texttt{cegpy} package models missingness in this way by default.  (Does stagedtrees also do this?)
However, including additional edges for missing values makes the event tree larger, increasing the complexity of the model and hindering the interpretability of the graphical representation.  When the missingness mechanism is not of direct interest (for example, when the data can be assumed to be MCAR), it seems preferable to not include it in the event tree to obtain a more parsimonious model. Because current learning algorithms assume a complete data set, this can be only achieved by either omitting any samples that contain missing values or manually imputing the missing values.  Omitting samples with missing values results in a waste of potentially useful data, while imputation can lead to artificially inflated confidence in the results of the analysis. 
%Traditional missing values have been handled in three possible ways in staged trees.

%The first is to completely discard any observation having even one variable missing. Or, in order to avoid this, imputation was performed before model estimation so to not decrease the sample size.

%A second option is to consider observations for structural learning and parameter estimation at depths of the tree where they are fully observed. An observation is then discarded when the variable where it has a missing value is considered. This is implemented in the \texttt{stagedtrees} package. Such an approach is similar to the node-average likelihood of \citet{balov2013consistent} and \citet{bodewes2021learning}.

%Third, missingness is considered as an additional level of a variable and formally modeled within the staged tree. This is mostly relevant when there is a missing not at random pattern and the tree can inform to the reason why missingness is observed \citep{barclay2014chain}. One downside of this approach is that the size of the tree increases, with the possibility of additional observed zeros. Furthermore, under the assumption of missing completely at random and missing at random this approach becomes less useful.

%By default the \texttt{cegpy} package takes this third approach including additional 'missing' edges whenever there is a missing value within the data.  Alternatively, one can choose to discard any observations with missing values.

\section{Staged Trees Model Selection with Missing Values}
\label{sec:algo}

A first challenge of using staged trees with missing data is in writing the likelihood function.  When data is missing, this should generally be explicitly modelled and included within the likelihood.  We begin by considering a single sample from the event tree $x=(x_1,\dots,x_k)$ (i.e. the values observed on a unit going from the root to a leaf through a single root-to-leaf path in the case of fully observed transitions) and split $x$ into its observed values $x_o$ and missing values $x_m$. We also define the missingness indicators $M=(m_1,\dots,m_k)$, where $m_i$  takes value one if $x_i$ is observed and zero if it is missing. As standard, $M$ is assumed independent of $\theta$. The probability of observing $x$ given $\theta$  is 
\[
    P(x \mid \theta) = P(x_o, M \mid \theta) 
    = \sum_{x_m} P(x_o, x_m, M \mid \theta) 
    = \sum_{x_m} P(x_o, x_m \mid \theta) P(M \mid x_o, x_m).
\]
When the data is MCAR, $M$ is independent of both the observed and missing values. Hence we have $P(x \mid \theta) = P(M) \sum_{x_m} P(x_o, x_m \mid \theta)$. When the data is MAR, $M$ is only independent of the missing values and it similarly holds $P(x \mid \theta) = P(M \mid x_o) \sum_{x_m} P(x_o, x_m \mid \theta)$. In both cases the missingness probability does not depend on $x_m$ and can be considered a constant in the likelihood function of $\theta$ given $x$. In the case of MAR we have
\begin{equation}
 L(\theta \mid x) = P(x \mid \theta) 
 = P(M \mid x_o) \sum_{x_m} P(x_o, x_m \mid \theta) 
 \propto \sum_{x_m} P(x_o, x_m \mid \theta). \label{eq:Probx}
\end{equation}
For the remainder of the section we assume the data to be MAR or MCAR so that the likelihood of $\theta$ can be considered separately from the missingness probabilities.

When $x$ contains no missing values it corresponds to a single root-to-leaf path $\lambda \in \Lambda$ and so $\sum_{x_m} P(x_o, x_m \mid \theta) = P(x_o \mid \theta) = \theta_{\lambda}$. However, when $x$ contains missing values the likelihood in Equation (\ref{eq:Probx}) requires summing over all possible completions of $x_m$. Each of these possible completions corresponds to a root-to-leaf path, which we call the \textbf{possible paths} of $x$.

%For a data set with no missing values the likelihood function in Equation (\ref{eq:likelihood}) factorizes nicely because each sample $x$ corresponds to a single root-to-leaf path $\lambda \in \Lambda$ and so its contribution to the likelihood function is simply $\theta_{\lambda}$.  In this case, the likelihood can be further re-organized into the individual contributions of the stage probabilities as in Equation (\ref{eq:factstage}).
%The result is a function that factorises among the individual transition probabilities. 
%However, when $x$ contains missing values it does not necessarily correspond to a single root-to-leaf path, but to a number of them, one for each possible observed combination of missing values. % Instead, there are a number of root to leaf paths that $x$ might correspond to, if the missing values could be observed.
%We call these the \textbf{possible paths} of $x$.

\begin{definition}
    Let $x=(x_1,\dots,x_k)$ be a sample from an event tree $\mathcal{T}$, which may include missing values.  The \emph{possible paths} of $x$ is a set $\Lambda_x \subseteq \Lambda$ such that $\lambda = (v_0,v_1,\dots,v_{k}) \in \Lambda_x$ if and only if $\mathrm{lab}(v_{i-1},v_i) = x_i$ for all $i$ such that $x_i$ is not missing.
\end{definition}

As an example, suppose we observed that a patient did not enter the ICU and did not die, but the information about whether she was intubated is missing in the event tree in Figure \ref{fig:compst}. The possible paths are $\{(v_0,v_1,v_3,\cdot),(v_0,v_1,v_4,\cdot)\}$, where $\cdot$ means going along the edge labeled D = No since leaves are not numbered in the tree.

Using the possible paths of $x$, we can rewrite the sum in Equation (\ref{eq:Probx}) in terms of root-to-leaf path probabilities as $L(\theta \mid x) \propto \sum_{\lambda \in \Lambda_{x}} \theta_\lambda$. The likelihood of the full data set $D$ of independent samples is then simply the product over all samples:
\begin{equation}
\label{eq:new}
 L( \theta \mid D) \propto \prod_{x \in D} \sum_{\lambda \in \Lambda_{x}} \theta_\lambda.
\end{equation}
This can be simplified  by collecting together all samples that have the same set of possible paths.  If we write all sets of possible paths present in the data set by $\Lambda_1,\dots,\Lambda_K$  and the number of samples that have possible paths $\Lambda_i$ by $n_i$, then the likelihood function can be written as
\begin{equation}\label{eq:likelihoodmissing}
  L( \theta \mid D) \propto \prod_{i=1}^{K} \left(\sum_{\lambda \in \Lambda_i} \theta_\lambda \right)^{n_i}.
\end{equation}

The likelihood function for fully observed data in Equation (\ref{eq:factstage}) is easy to work with due to its factorisation in the individual transition probabilities.  However, this is clearly no longer the case for the likelihood with missing data in Equations (\ref{eq:new}) and (\ref{eq:likelihoodmissing}). Hence, quantities related to the likelihood, such as the MLE or properties of the posterior distribution, often cannot be found analytically. There are two obvious strategies to circumvent this by approximating the likelihood. The first is to use a \textit{pseudo-likelihood} - a function that is somehow close to the full likelihood, but is analytically simpler and with closed-form estimators. The second is to use an approximating algorithm which directly targets the full likelihood, for instance the \textit{EM algorithm}.

\subsection{Pseudo-likelihoods}

The simplest pseudo-likelihood one can think of is by simply omitting any samples with missing values.  This simplifies the likelihood in Equation (\ref{eq:likelihoodmissing}) by only considering singleton sets of possible paths.  If we suppose that $\Lambda_1,\dots,\Lambda_{K_1}$ are the singleton possible paths and write their single entries by $\lambda_1,\dots,\lambda_{K_1}$ respectively, then the \textit{omit} pseudo-likelihood is 
\begin{equation}\label{eq:pseudolikelihoodomit}
  L_{\mathrm{Om}}( \theta \mid D) = \prod_{i=1}^{K_1} {\theta_{\lambda_i}}^{n_i}
\end{equation}
This has the same form as the likelihood for fully-observed data and so is computationally as simple.  However, it removes all terms associated to the possible paths $\Lambda_{K_1+1},\dots,\Lambda_{K}$ and so can be a poor approximation, especially when many samples contain missing values.

The approach currently implemented in \texttt{stagedtrees} is based on the fact that for a sample $x=(x_1,\dots,x_k)$ in which the first missing value is $x_j$, all paths $\lambda \in \Lambda_x$ have the same first $j$ vertices and are therefore associated to the same transition probabilities.  To extend this to the full data set, suppose that every $\lambda = (v_0,v_1,\dots,v_k) \in \Lambda_i$ has the same $v_0,\dots,v_{j_i}$.  Then the probabilities $\theta_{v_0,v_1},\dots,\theta_{v_{j_i-1},v_{j_i}}$ are common to all terms in $\sum_{\lambda \in \Lambda_i} \theta_\lambda$.  Hence we approximate this sum by the product of these common probabilities to give the following pseudo-likelihood, which we refer to as the \textit{first-missing} pseudo-likelihood
\begin{equation}\label{eq:pseudolikelihoodfirstmissing}
  L_{\mathrm{FM}}( \theta \mid D) = \prod_{i=1}^{K} \left( \prod_{j=1}^{j_i}\theta_{v_{j-1},v_j} \right)^{n_i}  
\end{equation}
The first-missing pseudo-likelihood still factorises thus leading to simple computations.  It also omits less of the data than the omit likelihood and so might be expected to better approximate the full likelihood.

 Utilising the equality between transition probabilities in the same stage gives another pseudo-likelihood.  Essentially, any observed value that can be unambiguously associated to a single stage can be used to estimate that stage transition probability.  This is  akin to the node-average likelihood for Bayesian networks \citep{balov2013consistent,bodewes2021learning} and so we adopt the name \textit{stage-average} pseudo-likelihood.  For example, if for every $\lambda = (v_0,v_1,\dots,v_k) \in \Lambda_i$, $v_{j-1}$ is in the same stage and there is a common $\mathrm{lab}(v_{j-1},v_j)$, then the transition probability $\theta_{v_{j-1},v_j}$ is common to all paths in $\Lambda_i$.  Writing $I_i$ for the set of indices for which this holds in $\Lambda_i$, we write the stage-average likelihood as
\begin{equation}\label{eq:pseudolikelihoodstageaverage}
  L_{\mathrm{SA}}( \theta \mid D) = \prod_{i=1}^{K} \left( \prod_{j\in I_i}\theta_{v_{j-1},v_{j}} \right)^{n_i}.  
\end{equation}
Notice that the stage-average pseudo-likelihood contains all terms that appear in the first-missing pseudo-likelihood, but might also contain additional terms. In terms of generality $L_{\mathrm{Om}} \prec L_{\mathrm{FM}} \prec L_{\mathrm{SA}}$ and therefore in principle the stage-average approach is expected to better approximate the full likelihood function. However, we can notice the following:

\begin{itemize}
\item The expression of the omit pseudo-likelihood is the same irrespective of the underling event tree $\mathcal{T}$ and coloring of the situations. It takes a data set and drops all rows with missing values irrespective of the model. Conversely, the first-missing pseudo-likelihood depends on the underlying event tree $\mathcal{T}$, but not on the coloring. Finally, the stage-average pseudo-likelihood depends on both the coloring and the event tree - this means any two staged trees might be estimated over different sets of data where different observed values are discarded.
\item Assume a fixed event tree. The stage-average and first-missing pseudo-likelihoods coincide for the saturated model, where each situation has its own color. For the full independence model, where all situations are in the same stage, the stage-average likelihood coincides with the full likelihood, since all observed values can be used to estimate the model.

\item Model selection under the first-missing and stage-average pseudo-likelihoods must be performed with caution since common scoring functions assume a common data set for all models compared \citep[most notably the BIC,][]{cohen2021normalized}. However, as noticed, these two pseudo-likelihoods might use different data sets to estimate the model. We will provide further comments on this issue in the discussion.

\item The stage-average likelihood is not further considered, because its implementation is challenging and would require a major update of the available software. This is because tables of observed counts for each situation must be constructed individually for each model considered. Again, more comments on this are provided in the discussion.
\end{itemize}

\subsection{The EM algorithm}

The EM algorithm is a popular computational tool for approximating the MLE or maximum a posteriori estimate in the presence of missing data \citep{dempster1977maximum}. In the context of probabilistic graphical models it was first introduced by \citet{lauritzen1995algorithm}. There are a number of proposed forms of EM algorithm, but the most common is for parameter estimation under a fixed model, which alternately updates the expected sufficient statistics (E step) and maximises parameter values (M step) until convergence.  For staged trees, the EM algorithm is initialised with some initial transition probabilities $\theta^{(0)}$.  Then the following two steps are iteratively applied
\begin{itemize}
    \item E step - calculate the expected path counts $n_\lambda^{(t)}$ given the data and current transition probabilities $\theta^{(t-1)}$.
    \item M step - calculate the maximised transition probabilities $\theta^{(t)}$ given the path counts $n_\lambda^{(t)}$.
\end{itemize}
The E step can be performed  using the possible paths of each sample where each sample is distributed among its possible paths according to the current transition probabilities.  That is, for a sample $x$ with possible paths $\Lambda_x$ and current transition probabilities $\theta^{(t-1)}$, if $\lambda \in \Lambda_x$ then the probability of $x$ following the path $\lambda$ is $\frac{\theta^{(t-1)}_\lambda}{\sum_{\lambda' \in \Lambda_x} \theta^{(t-1)}_{\lambda'}}$.  If $\lambda \not\in \Lambda_x$ then the probability is equal to 0.  Summing over all samples we get
\begin{equation}
\label{eq:suff}n_\lambda^{(t)} = \sum_{x: \lambda \in \Lambda_x} \frac{\theta^{(t-1)}_\lambda}{\sum_{\lambda' \in \Lambda_x} \theta^{(t-1)}_{\lambda'}}.
\end{equation}
 The M step is straightforward since it is analogous to finding the MLE for a complete data set - the only difference is that the expected path counts are not necessarily integers, but this does not change the calculation.  

The computation of the sufficient statistics $n_{\lambda}^{(t)}$ is expensive since it consists of multiple nested summations, as already noticed for BNs \citep{friedman1997learning}. For this reason, in practice a hard version of the EM algorithm is most often implemented where the computation of the sufficient statistics is replaced by direct imputation of all missing values in the data using the current transition probabilities
\citep[see e.g.][]{franzin2017bnstruct}. For staged trees, imputation of the missing values in a sample $x$ with current transition probabilities $\theta^{(t-1)}$ uses the probability $P(x_m \mid x_o, \theta^{(t-1)})$, which are equal to the conditional path probabilities in Equation (\ref{eq:suff}) \citep{thwaites2008propagation}. Although hard EM, consisting of single imputations, is known to be problematic \citep{schafer1999multiple}, it has been shown to have competitive performance, if not outperforming, standard EM in learning BNs \citep{ruggieri2020hard}. We henceforth only consider hard-EM algorithms.

%The hard EM algorithm is an adaptation for when the expected sufficient statistics cannot be or are expensive to calculate.  In this case the E step is replaced by direct imputation of all missing values in the data set.  This imputation is carried out according to the current transition probabilities.

The EM algorithm can be embedded within model selection by using the EM routine for each model estimated during the selection. However, this has been shown to be very computationally expensive. \citet{friedman1997learning,friedman1998bayesian} introduced the \textit{structural EM} algorithm for BNs which alternates E and M steps as in the traditional EM version: in the (hard) E-step, the data  is completed by imputation using the current model; in the M step, the model maximizing a model score (e.g. BIC) is identified using the complete data. The steps are repeated until there is no change in the model structure or a maximum number of iterations is reached. We adopt the same strategy to learn staged trees given a fixed event tree. For the M step of the structural EM, any of the currently available algorithms for learning staged trees could be used. We propose three possible strategies: using the backward hill-climbing algorithm (only merging stages) starting from the saturated model at each M step; using the hill-climbing algorithm (both merging and splitting stages at each iteration) starting from the saturated model at each M step; and using a hill-climbing algorithm starting from the model obtained at the previous M iteration.

The structural EM algorithm for staged trees can also be embedded within the dynamic programming approach which additionally learns the underlying event tree. However, as the experiments below demonstrate, this can become computationally expensive for larger event trees and model search algorithms that compare less models (such as the backward hill-climbing) should be preferred. 
%.  This either means it becomes computationally impractical, or model search algorithms that compare less models (such as the backward hill-climbing) should be preferred. 

%Either the standard or hard EM algorithm can be embedded within a model selection algorithm - this is referred to as EM within model selection.  This simply uses the EM algorithm to approximate the model score whenever it is needed in the model selection algorithm.  However, this approach can be computationally expensive, especially when the model selection algorithm compares many different models (since the EM algorithm must be run for all proposed models).  Instead, model selection within EM, or structural EM (SEM) can be used.  SEM alternates between finding the expected sufficient statistics (or imputing the data in the case of hard EM) given the current model and using a model selection algorithm given the sufficient statistics.  

\section{Experiments}

 We conduct a simulation study to evaluate the quality of the proposed approaches for selecting staged trees from data with missing values. The experiment was designed following the steps of \citet{ruggieri2020hard}. Data was simulated from five staged trees from the literature: Titanic \citep{carli2022r}, CHDS \citep{barclay2013refining}, bank advertising \citep{leonelli2024structural}, life quality \citep{varando2024staged}, and coronary \citep{leonelli2024structural}. Although the algorithms have been discussed for generic staged trees, here we consider only $X$-compatible staged trees to take full advantage of the capabilities of the \texttt{stagedtrees} R package. Details about these staged trees are given in Table \ref{tab:simu}.
\begin{table}
\floatconts
  {tab:simu}
  {\caption{Details of the staged trees considered in the simulation study.}}
  {\begin{tabular}{lccc}
  \toprule
	Staged Trees & \# Variables & \# Root-to-Leaf Paths & \# Stages \\
	\midrule
	Titanic & 4 & 32 & 13 \\
	CHDS & 4 & 24 & 7 \\
    Bank advertising & 4 & 16 & 8 \\
    Life quality & 5 & 72 & 17 \\
    Coronary & 6 & 64 & 14\\
\bottomrule
  \end{tabular}}
\end{table}

We controlled each of the following experimental conditions: missingness proportion ($p=0.05,0.10,0.20$), sample size ($N=500,1000,2500,5000$),  and missingness mechanism (MCAR, MAR, MNAR). 
%For an experiment a complete data set of size $N$ was simulated from a given staged tree.
A proportion $p$ of observed values from a complete data set of size $N$ was set as missing according to one of the three mechanisms using the \texttt{ampute} function of the \texttt{mice} R package \citep{schouten2018generating}. The default setup of \texttt{ampute} was used which specifies at most one missing entry per observation and equally splits the proportion of missing entries across variables. For each experimental condition the experiment was replicated 25 times.

We considered both the problem of learning the staging given a fixed event tree, and also the learning of the event tree. For the first case, we considered 9 algorithms: hill-climbing  and backward hill-climbing  using the complete data set (Full-HC and Full-BHC), the omit pseudo-likelihood (Om-HC and Om-BHC), the first-missing pseudo-likelihood (FM-HC and FM-BHC), the structural EM algorithm starting from the saturated model at each iteration (EM-HC and EM-BHC), and the EM algorithm with hill-climbing starting from the previously estimated model at each iteration  (EM-Simple). Models are selected using the BIC scoring rule. For the second case in which the event tree is also learned, the ninth approach is not considered. Due to slow computation related to the size of the event tree, experiments for learning the event tree are not carried out for the coronary staged tree, and for the life quality staged tree only the $N=500$ case is investigated.

To assess model selection, the normalized Hamming distance between the selected staged tree and the data generating staged tree is used. To assess probability estimation, and therefore predictive ability, both the Kullback-Leibler (KL) divergence and the Chan-Darwiche (CD) distance between the estimated and data generating root-to-leaf path probabilities are considered. The  time taken by each method to select and estimate the model is also measured. For routines that also estimate the underlying event tree, the Hamming distance is replaced by the Kendall distance between the true and selected variable orderings. 

We now summarize the results of the experiments. Selected plots are reported in the supplementary material. Of course, all measures of fit improve when the data set size increases, while learning time increases only slightly for larger data set sizes (since frequency tables are constructed once at the beginning of the algorithm). Furthermore, the performance of the algorithms is very similar for smaller data set sizes, while patterns become more apparent for larger ones (see e.g. Figure \ref{fig:titanic-ciao}). Henceforth, we focus only on the $N= 5000$ case. In most settings, the performance of the BHC and HC algorithms were comparable. However, for some measures and generating staged trees, either every BHC algorithm outperformed its HC counterpart or vice versa (see e.g. Figures \ref{fig:life-ciao}-\ref{fig:life-ciao2}). Importantly, the relative performance of the different missing data methods is maintained across the two model search methods. We henceforth only consider the class of HC algorithms to explicitly focus on the differences between the approaches proposed in this paper to handle missingness.

We start with the measures of fit in experiments where only the staging is estimated. The missingness proportion and mechanism have little effect on the Hamming distance (Figure \ref{fig:coro-hamming}). All algorithms perform comparably, highlighting that the presence of missing values does not hinder the capability of retrieving the true staging. The difference between the KL divergence of the various algorithms become more evident for larger proportions and depends on the missingness mechanism  (Figure \ref{fig:bank-kl}). For MCAR, the EMs perform worse, while Omit and First-Missing are comparable to the use of the full data. For MAR, the Omit algorithm performs worse with a critical decrease of performance for higher proportions of missingness. EMs outperform both Omit and First-Missing. For MNAR, EMs perform worse, but all algorithms that do not use the full data are far from Full-HC.
Similar results are observed for the CD distance (Figure \ref{fig:bank-cd}), but for some generating staged trees the differences between the algorithms are minimal (e.g. Titanic \ref{fig:titanic-cd}).

We next consider the experiments where the underlying event tree is also learned. For data generated from staged trees over four variables the Kendall distance is similar across algorithms with no effect of missingness proportion and mechanism, possibly due to the small number of possible orderings (Figure \ref{fig:chds-kendall}). For data simulated from the life quality staged tree we observe inconclusive patterns, where in some situations the Omit and First-Missing approach outperform Full-HC (Figure \ref{fig:life-kendall}). This is possibly due to the small data set size. Concerning the KL divergence and CD distance, we observe patterns similar to the experiment with a fixed event tree, where now the First-Missing and EM algorithms perform considerably worse than the Omit, with the exception of MAR missingness mechanisms (Figures \ref{fig:titanic-kll}-\ref{fig:bank-cdd})

Last, we consider learning times of our routines. For the most complex data generating staged tree (coronary), the Full-HC, Omit-HC, FM-HC and EM-Simple take almost the same time, with the EM-HC taking twice the time (just below one minute) (Figure \ref{fig:coronary-time}). The learning time of EM-HC slightly increases by missingness proportion and is overall slightly faster in the case of MCAR missingness. The learning time of EM-HC also shows more variability. 

For the experiments where we also estimate the event tree, it can be seen that the FM-HC is faster than Om-HC, which is in turn faster than Full-HC (Figure \ref{fig:chds-time1}). For the life quality staged tree, the EM-HC takes around five minutes and is considerably slower than the other approaches (Figure \ref{fig:life-time1}). In comparison, the EM-BHC algorithm only takes 8 seconds on average and is thus much faster than its HC counterpart, as observed for learning algorithms with no missing data \citep{carli2022r}. In all cases, missingness proportion has an effect on learning time, but missingness mechanism does not.

\section{Discussion}

This paper introduced a methodological formalization of missing values in staged trees and several approaches to account for them during model selection. The experimental study showed that the missingness mechanism and proportion have an effect on some measures of fit, but not on others. Depending on the missingness mechanism and underlying staged tree, different approaches might perform better than others. In terms of processing time, EM algorithms are not so distant from those over the full data set or based on pseudo-likelihoods.

Although the experimental study is rather comprehensive, it could be further extended. First, additional underlying staged trees could be considered. Second, algorithms that learn the event tree could be further investigated by only considering BHC algorithms which have been shown to be much faster. Third, additional ways to include missing values could be considered: for instance, by allowing more than one missing value per observation or by including missing values either close to the root or to the leaves of the tree.

In some of the experimental studies EM performed worse than the pseudo-likelihood methods, and in some the first-missing pseudo-likelihood performed worse than the omit. This initially seems counter intuitive because EM aims to utilise all observed data while first-missing utilises more of the data than omit - usually one would assume that more data leads to improved performance. One reason for this might be the use of BIC as scoring function in the model selection procedure. The BIC is known to have consistent model selection for staged trees with complete data sets \citep{gorgen2022curved}. This result extends to the omit pseudo-likelihood, as long as the number of fully observed samples tends to infinity as the whole sample size increases.

However, consistency does not extend to the first-missing pseudo-likelihood or EM - the BIC is in fact known to not be consistent for BNs with missing data \citep{balov2013consistent}. 
%and to have issues more generally when there is missing data \citep{cohen2021normalized}.
In particular, a data set with missing values contains strictly less information than the same fully observed data set. This means that the rate of convergence of the estimated transition probabilities is slower when there is missing data. It stands to reason that the penalty term in the BIC should be changed to reflect this slower rate of convergence. More research is required to find an adaptation of the BIC 
%or other scoring function
that is more appropriate in missing data settings for staged trees. Implementing such a scoring function will require a significant update of \texttt{stagedtrees}, which currently uses the standard \texttt{BIC} method for \texttt{logLik} objects, that have a single value for the data set size and number of parameters. Pseudo-likelihoods and EM require the implementation of a tailored \texttt{BIC} method, just as  in \texttt{bnlearn} \citep{scutari2010learning}.

%The experimental studies showed that in specific cases, especially when also learning the event tree, the first-missing algorithms performed worst. This might be due to the use of the BIC as model selection measure. \citet{cohen2021normalized} and \citet{balov2013consistent} demonstrated that the BIC is not consistent and that a stronger penalization must be considered. Future work will focus on identifying consistent model selection criteria for first-missing and stage-average pseudo-likelihoods. It is expected that results similar to those of BNs will hold, since \citet{gorgen2022curved} proved that the BIC is consistent for complete data sets just as BNs. Future work will also focus on implementing the computation of such model selection criteria with missing values. This will require a significant update of \texttt{stagedtrees}, which currently uses the standard \texttt{BIC} method for \texttt{logLik} objects, that have a single number of data set size and free parameters. Therefore,  pseudo-likelihoods require the implementation of a tailored \texttt{BIC} method, just as implemented in \texttt{bnlearn} \citep{scutari2010learning}.  

The \texttt{stagedtrees} package now includes an implementation of the hard-EM algorithm. However, the experiments show that its performance is variable and that in several instances pseudo-likelihoods gave better results. It is possible that traditional or soft EM approaches might be better suited for staged trees and provide a significant improvement on the results. We plan to implement and evaluate soft-EM approaches in future research.

Lastly, an implementation and performance investigation of the stage-average likelihood is planned for future research, but, again, this requires a major update of the \texttt{stagedtrees} R package. Currently, \texttt{stagedtrees} constructs the tables of observed counts for each situation only once at the beginning of the algorithm. While this approach is still compatible with the first-missing pseudo-likelihood, it is not with the stage-average one, where the tables must be constructed every time a new model is considered during the search. The stage-average likelihood is expected to outperform the other pseudo-likelihood methods since it is the closest to the full likelihood.

\bibliography{references}

\appendix
\newpage
\section{Results of the Experiments}\label{apd:first}

\begin{figure}[htbp]
 % Caption and label go in the first argument and the figure contents
 % go in the second argument
 % the third argument is the figure; without subfigure just use \include... directly
\floatconts
  {fig:titanic-ciao}
  {\caption{KL divergence for data simulated from the titanic staged tree by missingness proportion and data set size (combining all missingness mechanisms, only HC algorithms).}}
  {%
  \includegraphics[width=0.82\linewidth]{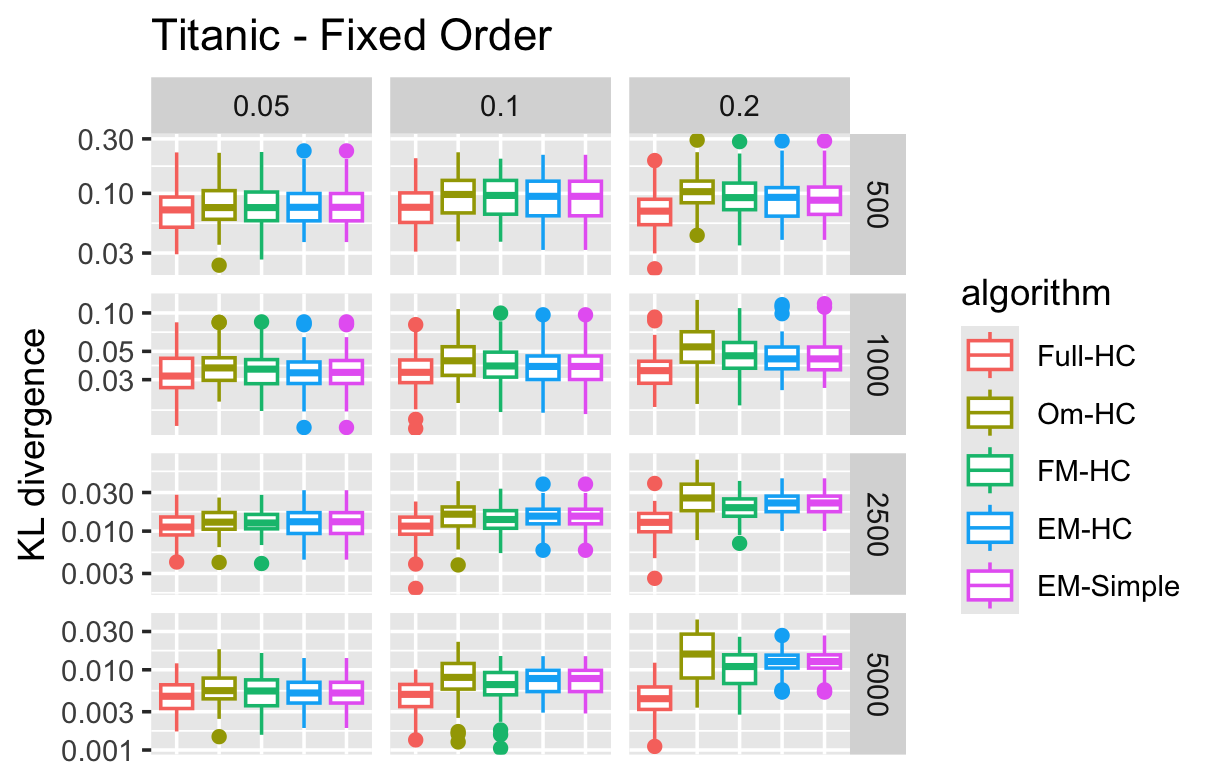}
  }
\end{figure}

\begin{figure}[htbp]
 % Caption and label go in the first argument and the figure contents
 % go in the second argument
 % the third argument is the figure; without subfigure just use \include... directly
\floatconts
  {fig:life-ciao}
  {\caption{KL divergence for N = 5000 data simulated from the life quality staged tree by missingness proportion and mechanism.}}
  {%
  \includegraphics[width=0.82\linewidth]{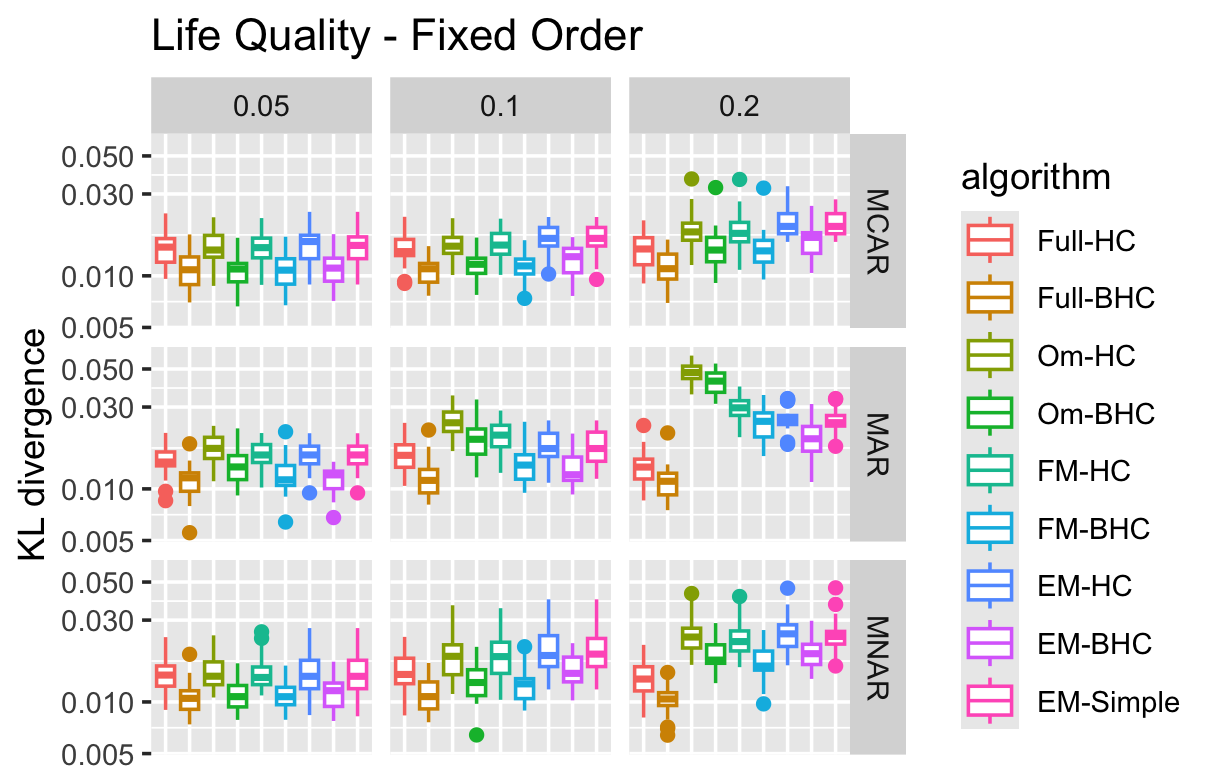}
  }
\end{figure}

\begin{figure}[htbp]
 % Caption and label go in the first argument and the figure contents
 % go in the second argument
 % the third argument is the figure; without subfigure just use \include... directly
\floatconts
  {fig:life-ciao2}
  {\caption{Normalized Hamming distance for N = 5000 data simulated from the life quality staged tree by missingness proportion and mechanism.}}
  {%
  \includegraphics[width=0.82\linewidth]{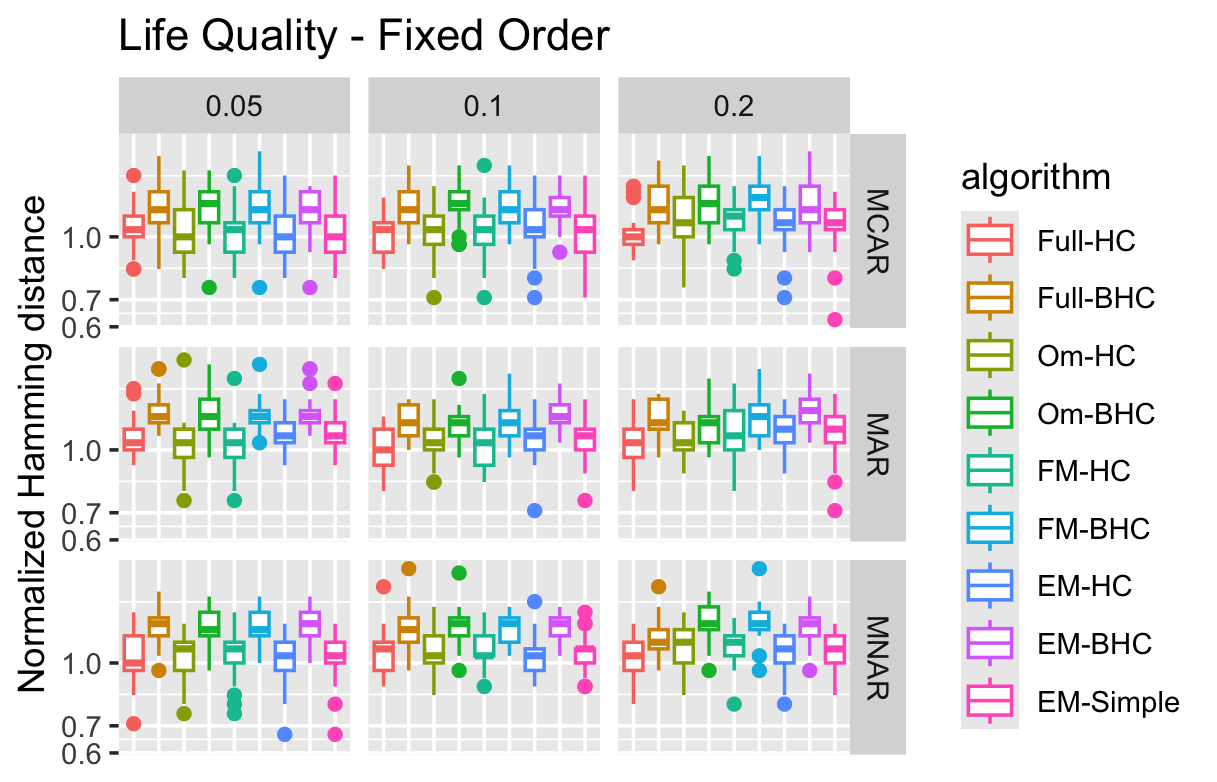}
  }
\end{figure}

\begin{figure}[htbp]
 % Caption and label go in the first argument and the figure contents
 % go in the second argument
 % the third argument is the figure; without subfigure just use \include... directly
\floatconts
  {fig:coro-hamming}
  {\caption{Normalized Hamming distance for N = 5000 data simulated from the coronary staged tree by missingness proportion and mechanism.}}
  {%
  \includegraphics[width=0.82\linewidth]{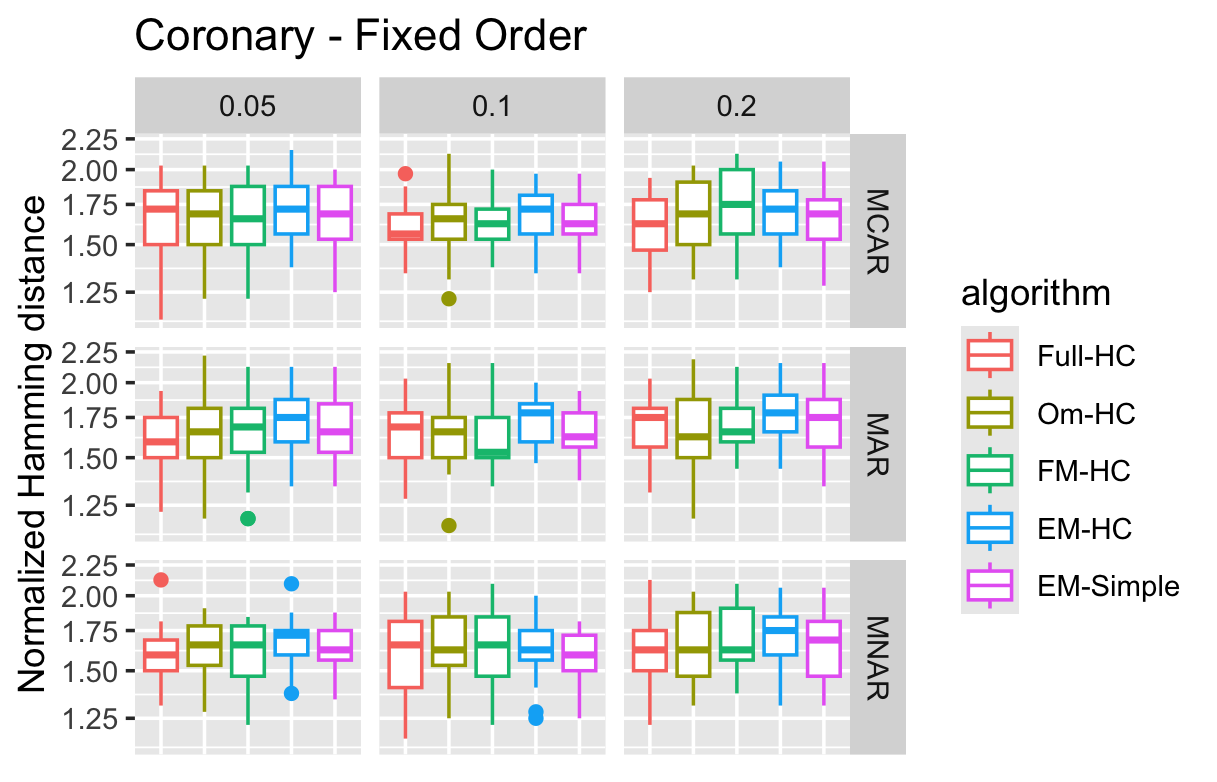}
  }
\end{figure}

\begin{figure}[htbp]
 % Caption and label go in the first argument and the figure contents
 % go in the second argument
 % the third argument is the figure; without subfigure just use \include... directly
\floatconts
  {fig:bank-kl}
  {\caption{KL divergence for N = 5000 data simulated from the bank staged tree  by missingness proportion and mechanism.}}
  {%
  \includegraphics[width=0.82\linewidth]{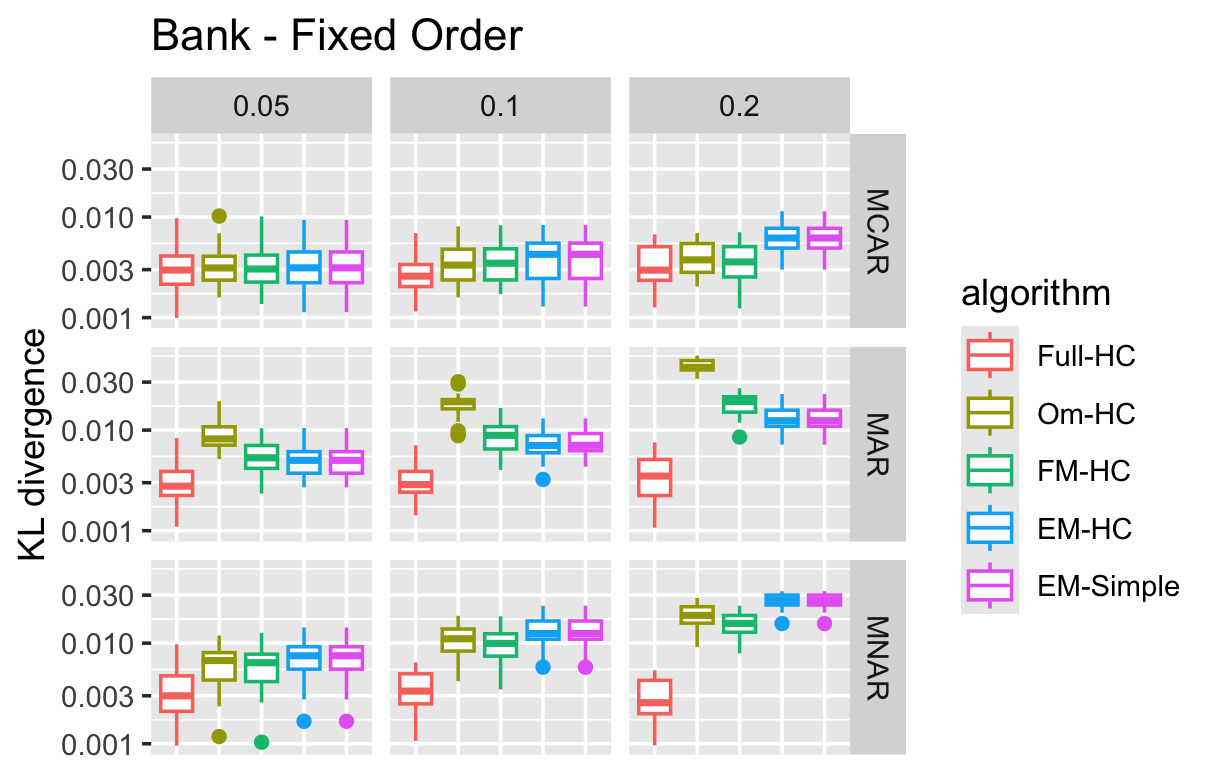}
  }
\end{figure}

\begin{figure}[htbp]
 % Caption and label go in the first argument and the figure contents
 % go in the second argument
 % the third argument is the figure; without subfigure just use \include... directly
\floatconts
  {fig:bank-cd}
  {\caption{CD distance for N = 5000 data simulated from the bank staged tree by missingness proportion and mechanism.}}
  {%
  \includegraphics[width=0.82\linewidth]{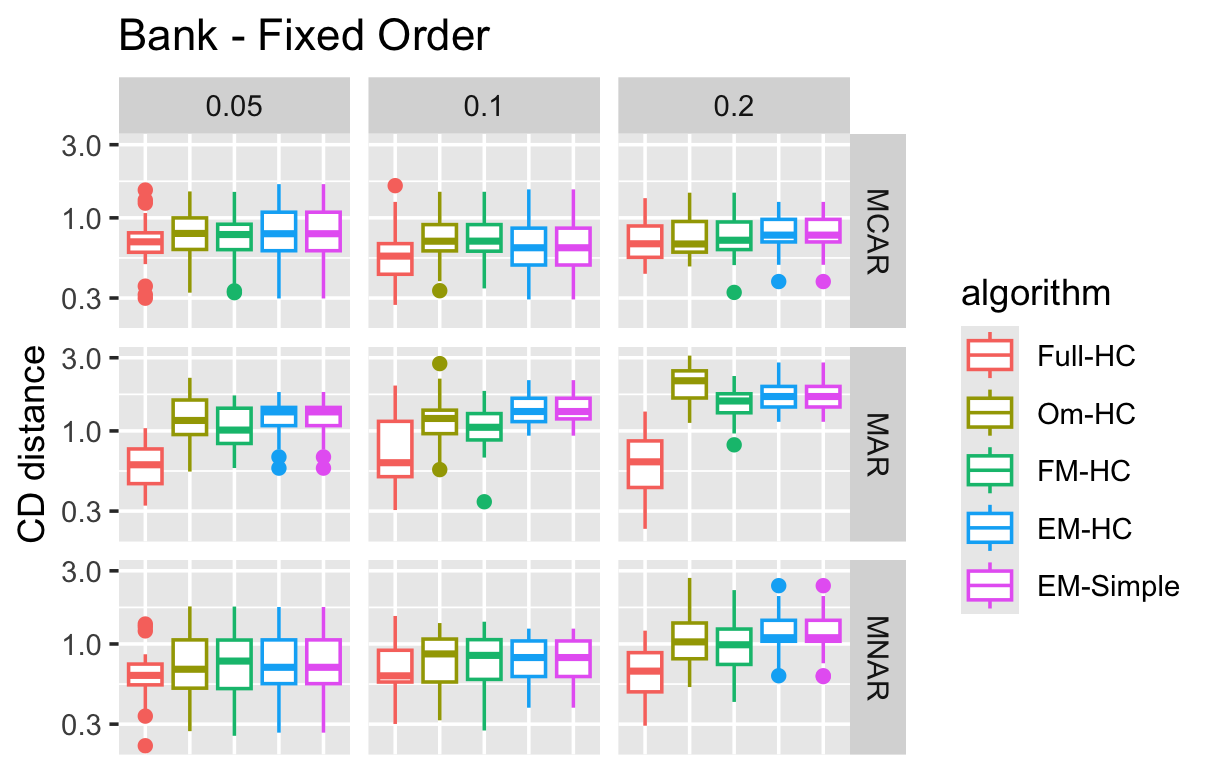}
  }
\end{figure}

\begin{figure}[htbp]
 % Caption and label go in the first argument and the figure contents
 % go in the second argument
 % the third argument is the figure; without subfigure just use \include... directly
\floatconts
  {fig:titanic-cd}
  {\caption{CD distance for N = 5000 data simulated from the titanic staged tree by missingness proportion and mechanism.}}
  {%
  \includegraphics[width=0.82\linewidth]{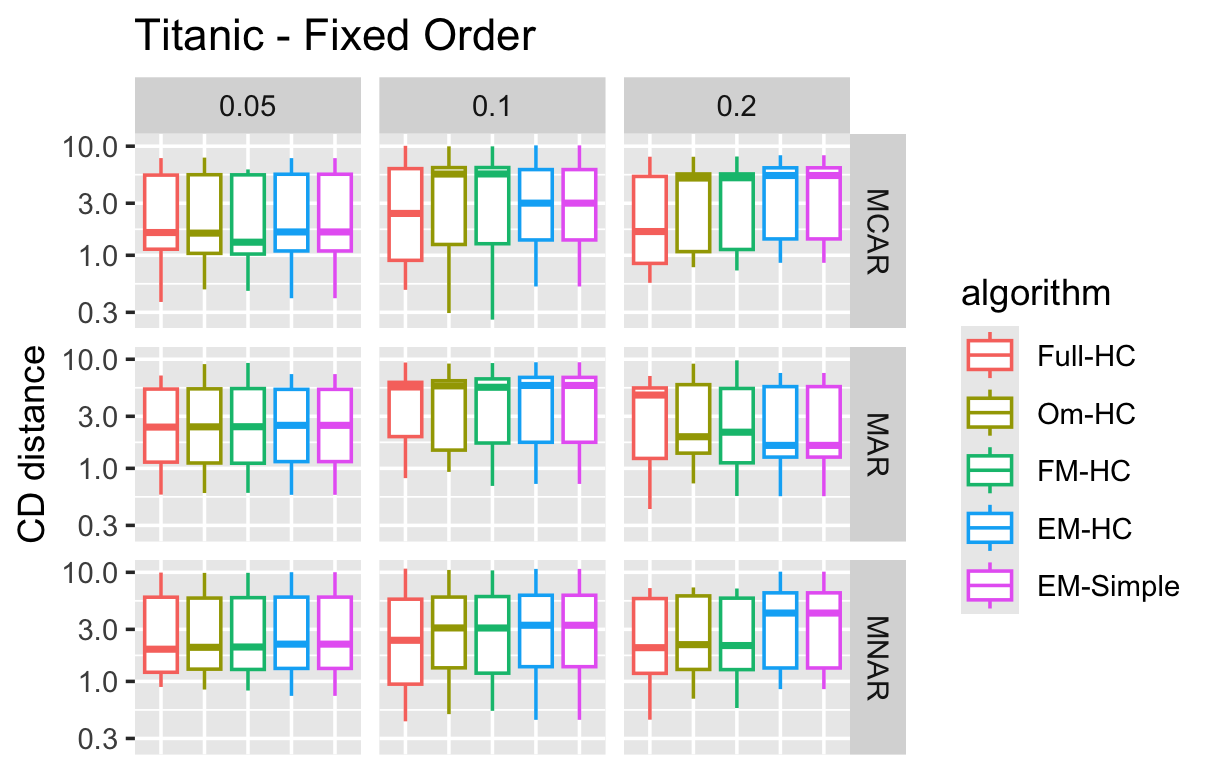}
  }
\end{figure}

\begin{figure}[htbp]
 % Caption and label go in the first argument and the figure contents
 % go in the second argument
 % the third argument is the figure; without subfigure just use \include... directly
\floatconts
  {fig:chds-kendall}
  {\caption{Kendall distance for N = 5000 data simulated from the chds staged tree by missingness proportion and mechanism.}}
  {%
  \includegraphics[width=0.82\linewidth]{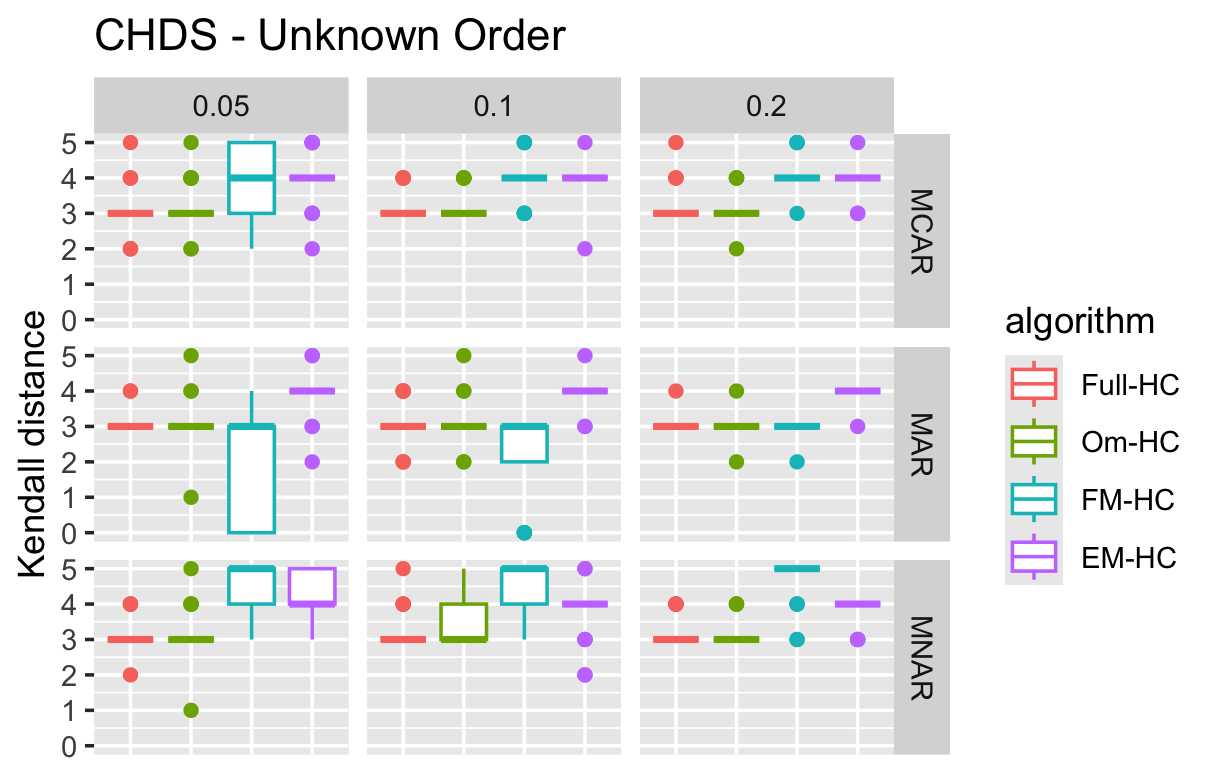}
  }
\end{figure}

\begin{figure}[htbp]
 % Caption and label go in the first argument and the figure contents
 % go in the second argument
 % the third argument is the figure; without subfigure just use \include... directly
\floatconts
  {fig:life-kendall}
  {\caption{Kendall distance for N = 500 data simulated from the life quality staged tree by missingness proportion and mechanism.}}
  {%
  \includegraphics[width=0.82\linewidth]{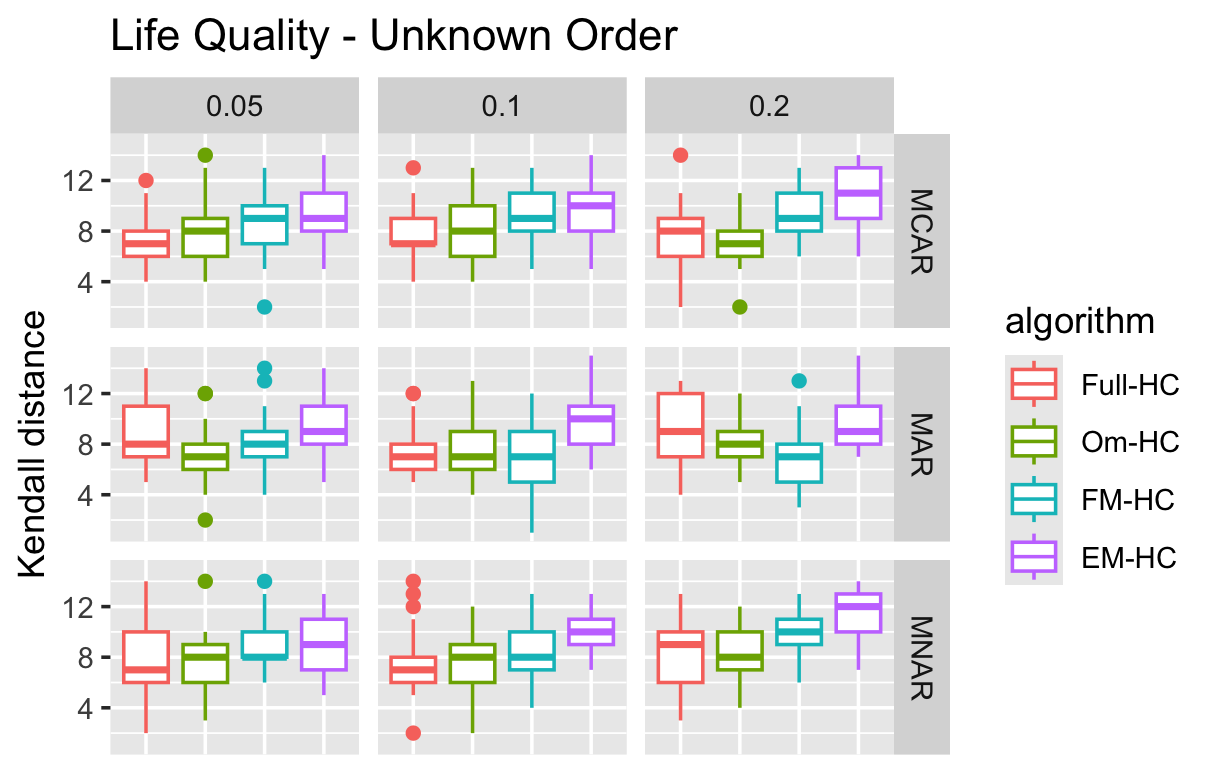}
  }
\end{figure}

\begin{figure}[htbp]
 % Caption and label go in the first argument and the figure contents
 % go in the second argument
 % the third argument is the figure; without subfigure just use \include... directly
\floatconts
  {fig:titanic-kll}
  {\caption{KL divergence for N = 5000 data simulated from the titanic staged tree by missingness proportion and mechanism.}}
  {%
  \includegraphics[width=0.82\linewidth]{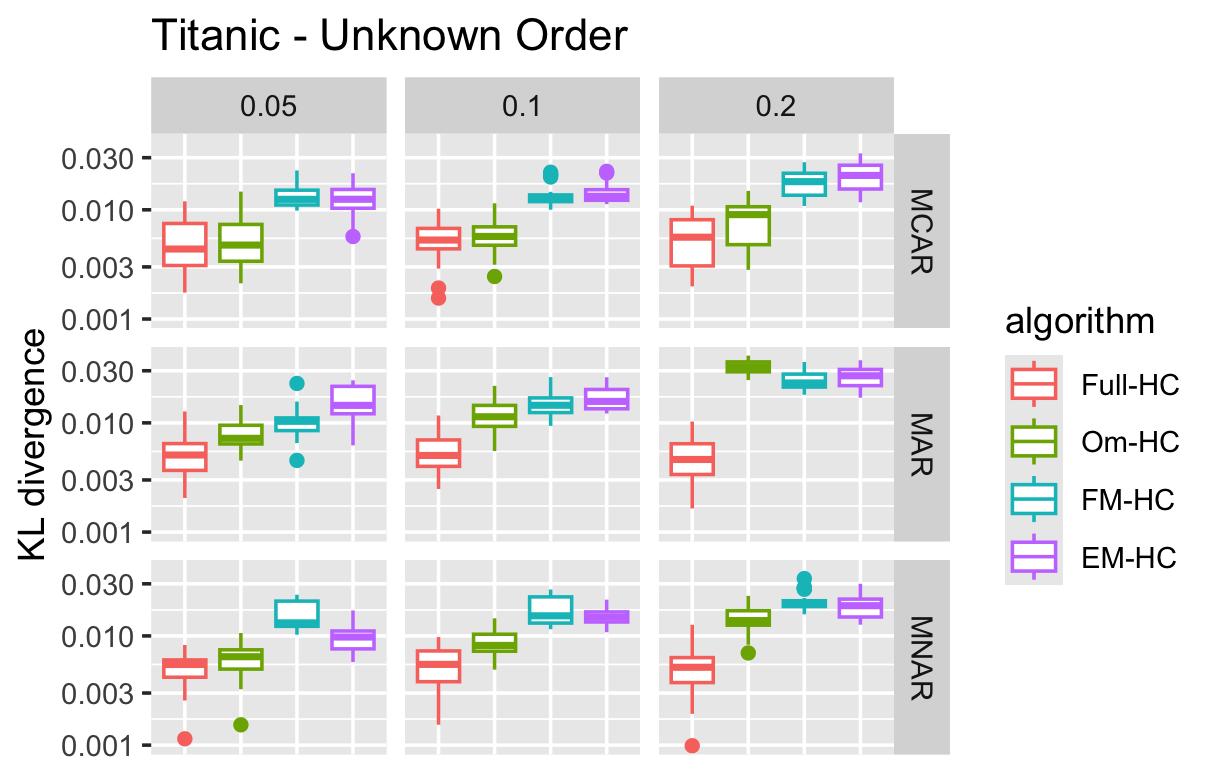}
  }
\end{figure}

\begin{figure}[htbp]
 % Caption and label go in the first argument and the figure contents
 % go in the second argument
 % the third argument is the figure; without subfigure just use \include... directly
\floatconts
  {fig:bank-cdd}
  {\caption{CD distance for N = 5000 data simulated from the bank staged tree by missingness proportion and mechanism.}}
  {%
  \includegraphics[width=0.82\linewidth]{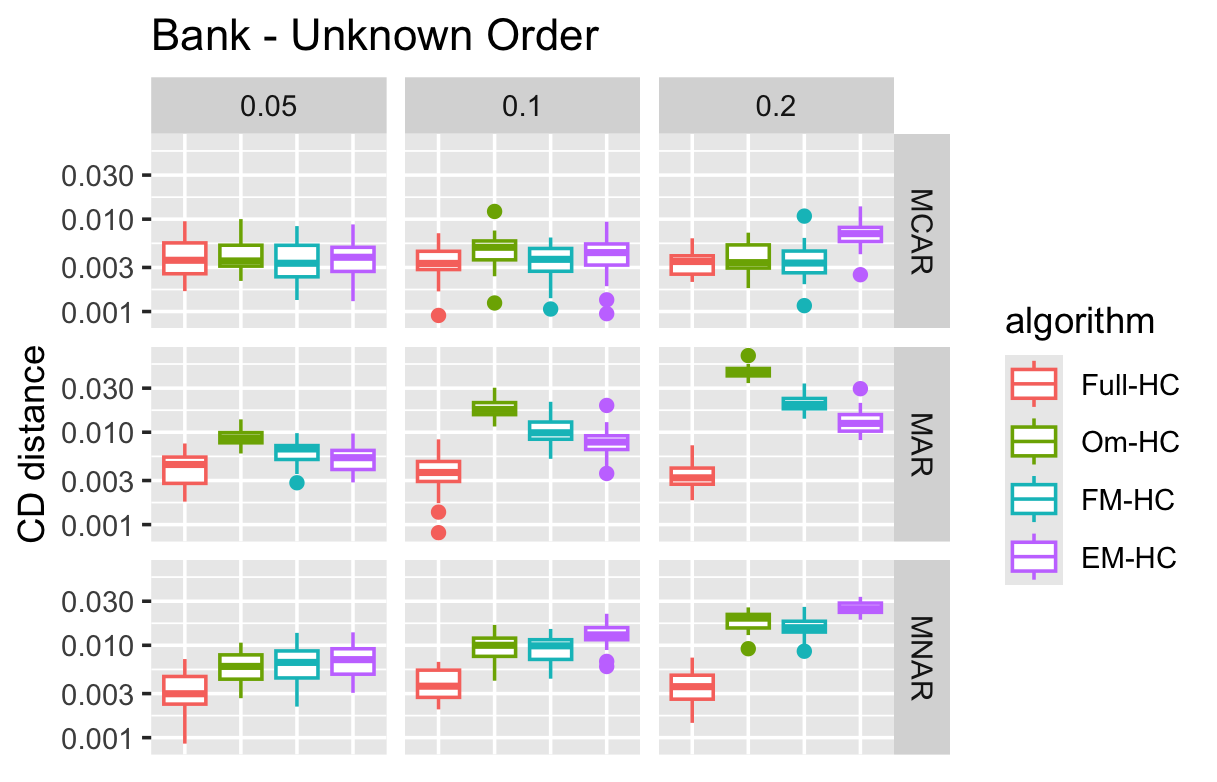}
  }
\end{figure}

\begin{figure}[htbp]
 % Caption and label go in the first argument and the figure contents
 % go in the second argument
 % the third argument is the figure; without subfigure just use \include... directly
\floatconts
  {fig:coronary-time}
  {\caption{Learning time for N=5000 data simulated from the coronary staged tree by missingness proportion and mechanism.}}
  {%
  \includegraphics[width=0.82\linewidth]{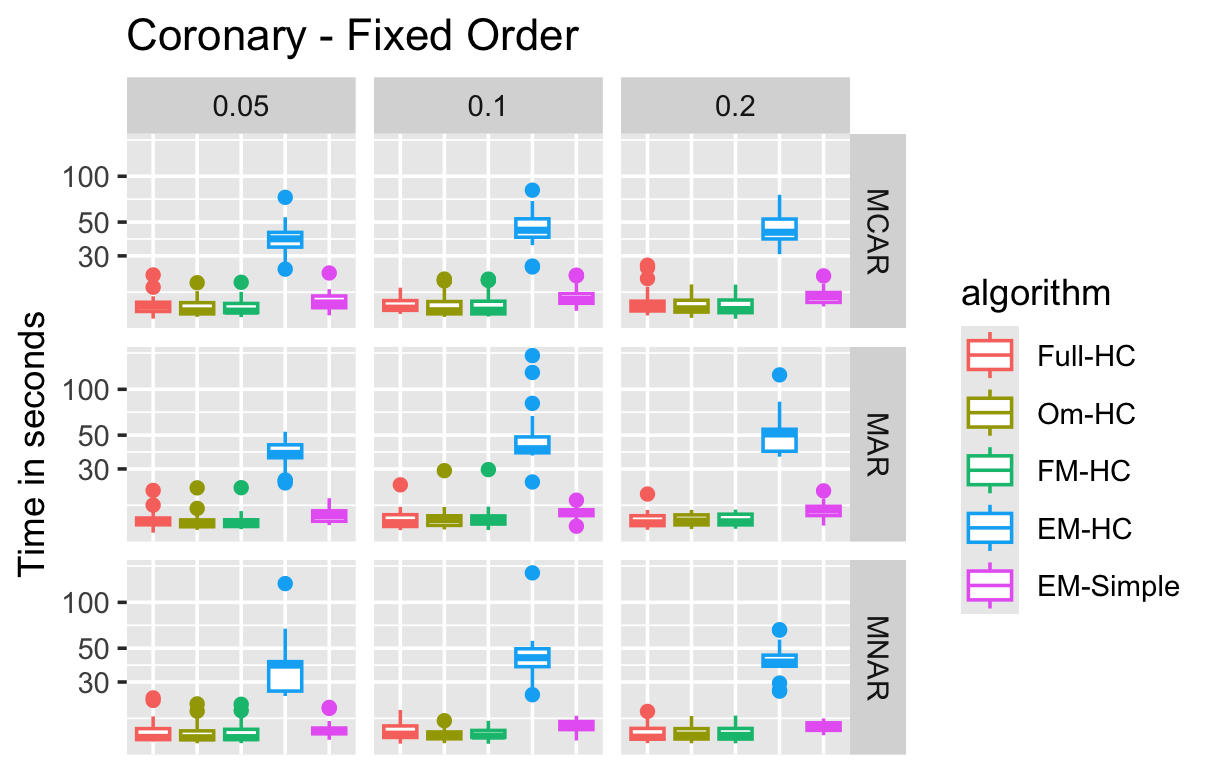}
  }
\end{figure}

\begin{figure}[htbp]
 % Caption and label go in the first argument and the figure contents
 % go in the second argument
 % the third argument is the figure; without subfigure just use \include... directly
\floatconts
  {fig:chds-time1}
  {\caption{Learning time for N=5000 data simulated from the chds staged tree by missingness proportion and mechanism.}}
  {%
  \includegraphics[width=0.82\linewidth]{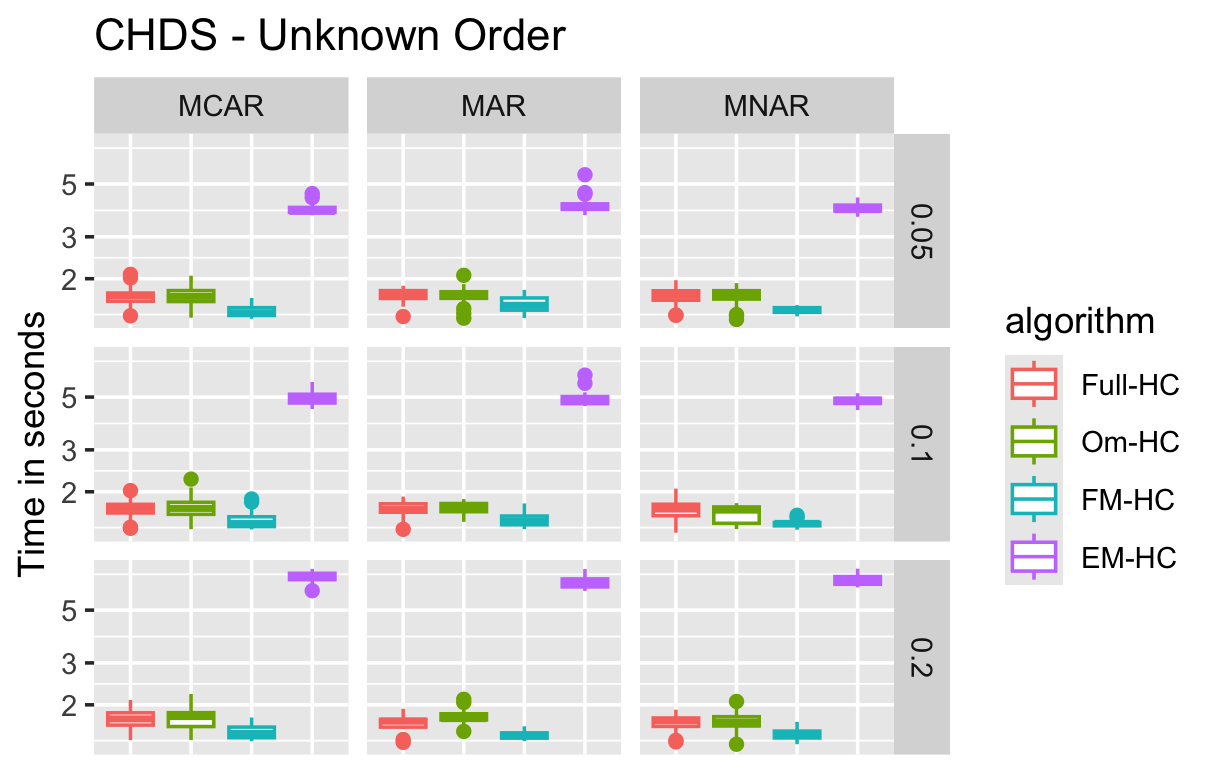}
  }
\end{figure}

\begin{figure}
 % Caption and label go in the first argument and the figure contents
 % go in the second argument
 % the third argument is the figure; without subfigure just use \include... directly
\floatconts
  {fig:life-time1}
  {\caption{Learning time for N=500 data simulated from the life quality staged tree by missingness proportion and mechanism.}}
  {%
  \includegraphics[width=0.82\linewidth]{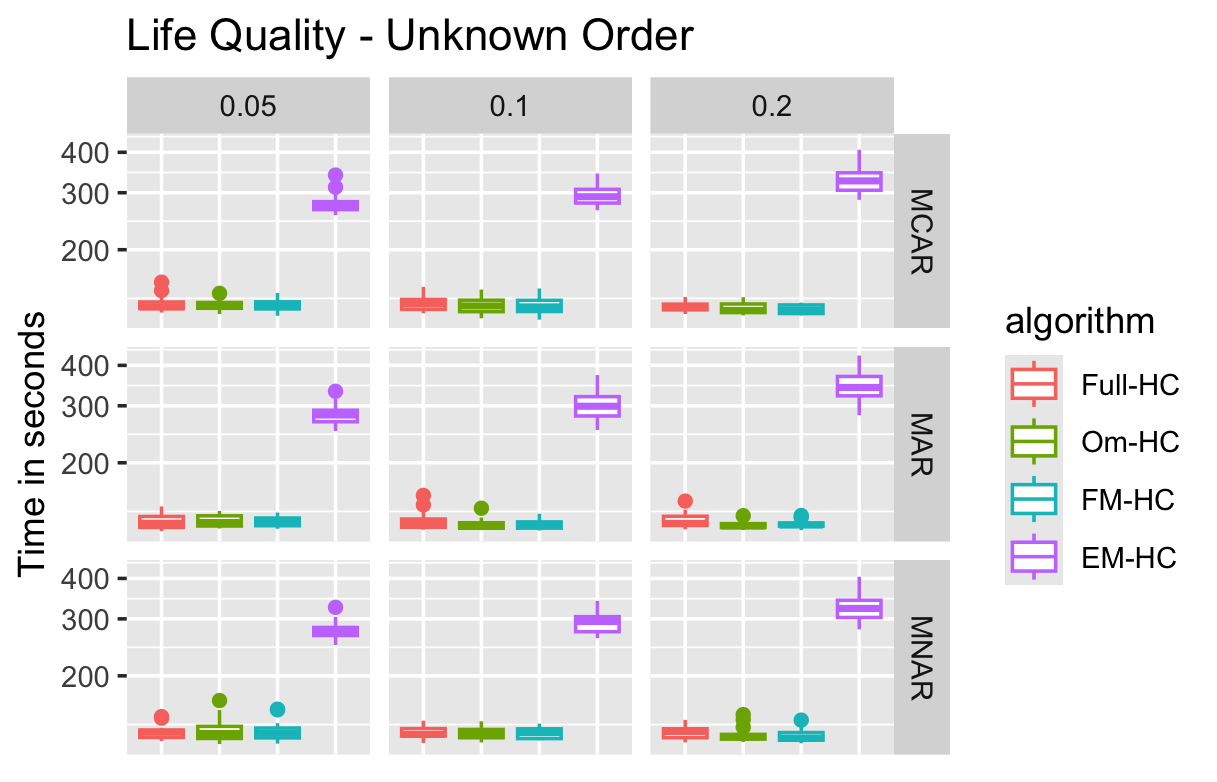}
  }
\end{figure}

\end{document}